%% file: iclr_2027.tex
\documentclass{article}

\usepackage{iclr2027_conference,times}

\input{preamble}

\input{titleblock}

\iclrfinalcopy

\begin{document}

\maketitle
\lhead{Preprint}

\begin{abstract}
\input{abstract}
\end{abstract}

\input{body}

\section*{Reproducibility statement}

Unless labelled otherwise, every success rate is a mean over 50 episodes \(\times\) 5
evaluation seeds \(\times\) 3 training seeds, reported with the population
standard deviation across the three training-seed means; the protocol,
including the fixed 50-step episode budget, is stated in
\S\ref{sec:exp-setup} and Tab.~\ref{tab:lewm-main}, and every cell that rests on a single training seed
is labelled as such where it appears. Environments, datasets and
their collection oracles are described in App.~\ref{app:envs}; optimiser and
training-time hyperparameters in App.~\ref{app:optim}; the planner's
cross-entropy-method and receding-horizon settings, which govern every
goal-conditioned number, in App.~\ref{app:cem-mpc-config}. Every baseline we
compare against under a matched protocol (LeWM and PLDM on all four
environments, DINO-WM on Cube and Cube-Double) was retrained in our own
pipeline rather than transcribed, so the data, training scaffold and
evaluation budget are shared across those rows
(Tab.~\ref{tab:lewm-main}, \S\ref{sec:exp-main}).
The training and evaluation code, every configuration file behind the
reported runs, and the pinned environment lockfile are available at
\url{https://github.com/corl-team/qantara}.

\section*{AI use statement}

We used AI assistants in two ways. For research execution, AI coding
assistants helped write training and evaluation code and make
figures; the authors set the research direction, reviewed and
tested all code, and recomputed every reported number from the stored
run outputs. For writing, AI assistants helped edit and polish
author-written text. AI tools were not used to generate data, and the
method and its derivations are the authors' own. The authors take full
responsibility for the content of this paper.

\bibliographystyle{iclr2027_conference}
\bibliography{references}

\input{appendix}

\end{document}

%% file: preamble.tex
\usepackage[utf8]{inputenc}    %
\usepackage[T1]{fontenc}       %
\usepackage{microtype}         %

\usepackage{amsmath}           %
\usepackage{amsfonts}          %
\usepackage{amssymb}           %
\usepackage{mathtools}         %
\usepackage{bm}                %
\usepackage{nicefrac}          %

\usepackage{graphicx}          %
\usepackage{xcolor}            %
\usepackage{subcaption}        %
\usepackage{wrapfig}            %
\usepackage{tikz}              %
\usetikzlibrary{arrows.meta}

\usepackage{booktabs}          %
\usepackage{array}             %
\usepackage{multirow}          %
\usepackage{makecell}          %
\usepackage{siunitx}           %
\usepackage{caption}           %

\usepackage{enumitem}          %

\usepackage{algorithm}         %
\usepackage{algpseudocode}     %

\usepackage{listings}          %
\usepackage{tcolorbox}         %

\definecolor{codebg}{rgb}{0.97,0.97,0.97}
\definecolor{codekw}{rgb}{0.0,0.4,0.8}
\definecolor{codecmt}{rgb}{0.4,0.55,0.4}
\definecolor{codestr}{rgb}{0.6,0.2,0.2}
\lstdefinestyle{pythonstyle}{
  language=Python,
  basicstyle=\ttfamily\footnotesize,
  keywordstyle=\color{codekw}\bfseries,
  commentstyle=\color{codecmt}\itshape,
  stringstyle=\color{codestr},
  showstringspaces=false,
  breaklines=true,
  frame=none,
  columns=fullflexible,
  morekeywords={randn_like, repeat_interleave},
}

\usepackage{hyperref}          %
\usepackage{url}               %
\usepackage[capitalize,noabbrev]{cleveref}  %

%% file: titleblock.tex
\title{Qantara: Bridge-Flow Training \\ for Multi-Paradigm JEPA Inference}

\author{%
  Ruslan Rakhimov\thanks{Correspondence: \texttt{ru.rakhimov@t-tech.dev}. Project page: \url{https://corl-team.github.io/qantara}} \quad
  George Bredis \quad Yuriy Maksyuta \quad Daniil Gavrilov \\
  T-Tech
}

%% file: abstract.tex
Joint-Embedding Predictive Architectures (JEPAs) underpin a growing family of latent world models for control from raw pixels. Published JEPA world models for pixel-based control (PLDM, DINO-WM, LeWM, V-JEPA-2-AC) each commit at training time to a single inference paradigm, trajectory optimisation in a learned dynamics model, so a checkpoint trained to plan cannot be read off as a policy without retraining. Deferring that choice to inference would let the constraints that bind at deployment, per-step latency and whether a goal observation can be supplied, pick the path. We present \textbf{Qantara}, an end-to-end JEPA of \(\sim\!21\)M parameters whose joint training objective pairs a Brownian-bridge interpolant between consecutive clean latents on the state axis with noise-to-data flow matching on the action axis. One checkpoint serves three inference paradigms without retraining: latent planning, behaviour-cloning action sampling, and a video--inverse composition that predicts the next latent without action conditioning, then recovers the action bridging the two. Training samples the two noise levels along the edges of the (action-noise, state-noise) square, one edge per inference dispatch, so each path is trained where it is queried. Replacing that sampler with uniform sampling over the square at matched compute drops Push-T planning from 90.1 to 53.3 success rate (SR). On the LeWM control suite at a matched 10-epoch budget, Qantara averages 91.2\,SR over the four environments and reaches 93.7 on OGBench-Cube, \(+7.7\) over DINO-WM and \(+17.0\) over LeWM, each measured against the stronger of that baseline's published score and our retrain of it. From the same weights the two goal-blind paths, which receive no goal image, reach 82.1--83.2\,SR on Push-T and 70.8--72.8 on Cube at \(\sim\!15\)--\(65\times\) lower inference cost, and fall to near-noise on Reacher, whose released training corpus comes from a random policy.

%% file: body.tex
\newcommand{\qtfit}[1]{\resizebox{\textwidth}{!}{#1}}

\section{Introduction}
\label{sec:intro}

A robot deployed for visuomotor control faces inference-time
constraints (per-step latency, the size of the action search
space, whether a goal observation can be supplied) that the
training run cannot anticipate. The same learned predictor is
useful in different ways under different constraints: rolled out
as a forward dynamics simulator under a planner when the search
budget allows, or read off as a behaviour-cloning policy when
latency is tight. Published JEPA world models trained from pixels
for control commit to one of these inference paradigms at
training time, and a checkpoint trained for one cannot serve the
others without retraining.

Two clusters of prior work tile the relevant design space.
JEPA world models commit to a single inference paradigm,
trajectory optimisation: PLDM \citep{sobal2025stresstesting},
DINO-WM \citep{zhou2024dinowm}, and LeWM
\citep{maes2025lewm,balestriero2025lejepa} train sub-billion
action-conditional latent dynamics and plan by minimising a
goal-conditioned latent distance, and V-JEPA-2-AC
\citep{assran2025vjepa2} post-trains a 300M action-conditional
predictor on a frozen 1B video JEPA and plans the same way, by
cross-entropy-method energy minimisation in a receding-horizon
loop. Each of them needs a goal image and a rollout budget at
every control step. Multi-paradigm checkpoints exist in a second
cluster that predicts the future observation rather than a
latent dynamics of its own, at up to billions of parameters:
PAD, UVA, UWM, DUST, and Cosmos Policy
\citep{guo2024pad,li2025uva,zhu2025uwm,won2025dust,kim2026cosmospolicy}
denoise observations and actions jointly and serve forward
dynamics, inverse dynamics, and behaviour policy from one set
of weights. The (sub-billion JEPA, multi-paradigm) cell is
sparsely occupied: \citet{nguyen2026gcidm} reach it with a second
training stage fitted on frozen LeWM latents.

The obstacle is that planning, behaviour cloning, and inverse
dynamics each query the predictor at a different combination of
clean and noisy inputs on the action and state axes, so an
objective tuned for one combination under-trains the others.
The open question is whether a single sub-billion JEPA can
support all three at deployment without a cost to any of them.

We present \textbf{Qantara}, an end-to-end pixel
JEPA world model whose joint training objective pairs a
Brownian-bridge interpolant on the state axis between
consecutive clean latents with noise-to-data flow matching on
the action axis
\citep{lyu2024framebridge,liu2023ipsb,lipman2022flow,albergo2023stochastic}.
Per-token noise levels \((\tau^a, \tau^z)\) are sampled along
the four edges of the \([0,1]^2\) noise square (one edge per
inference dispatch) plus a fifth diagonal mode that
co-regularises the shared trunk. The same training objective
populates the four corners that planning, behaviour cloning,
and inverse dynamics query at deployment. From a single
\(\sim\!21\)M-parameter checkpoint, Qantara serves three
inference paradigms without retraining: goal-conditioned
latent planning, behaviour-cloning sampling, and a
video--inverse composition that first predicts the next
latent without action conditioning and then recovers the
action that bridges the two.

We make three contributions.
(1) We present Qantara, a sub-billion JEPA world model that
defers the planning-vs-imitation choice to inference
(\S\ref{sec:method}).
(2) On the LeWM-suite~\citep{maes2025lewm} at a matched 10-epoch
budget, Qantara averages 91.2\,SR and reaches 93.7 on
OGBench-Cube, \(+7.7\) over DINO-WM as re-evaluated
by~\citet{maes2025lewm}; from the same checkpoint, the
behaviour-cloning and video--inverse dispatches read off at
\(\sim\!15\)--\(65\!\times\) lower inference cost than Qantara's CEM
planner (\S\ref{sec:exp-main}, \S\ref{sec:exp-bc}).
(3) We identify a corner co-regularisation effect: dropping any
of the \texttt{policy}, \texttt{inverse}, or \texttt{joint}
modes destabilises Push-T training on all three inference paths,
even the planner, whose edge shares at most one corner with theirs
(\S\ref{sec:exp-modes}).

\section{Method}
\label{sec:method}

\begin{figure}[t]
  \centering
  \includegraphics[width=\linewidth]{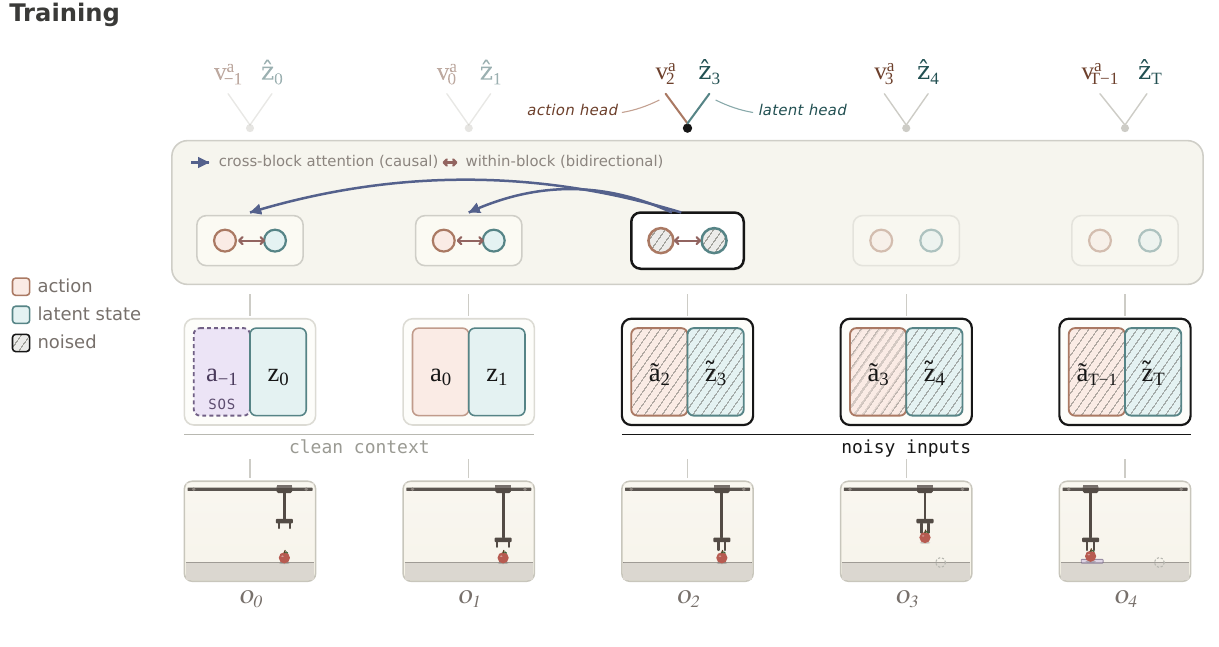}
  \caption{\textbf{Qantara training scheme.} Block-causal predictor
  on \((a, z)\) tokens; a clean prefix conditions the noisy
  continuation, arrows mark where attention is causal (across
  blocks) versus bidirectional (within a block), and per block the
  action head emits an action velocity \(v^a\)
  while the state head emits a latent residual that yields \(\hat z\).}
  \label{fig:qantara-modes}
\end{figure}

\textbf{Qantara} is a joint model of latents and their
generating actions, trained offline from trajectories
\(\{(o_t, a_t)\}_{t=0}^{T}\) of pixel observations \(o_t\) and
actions \(a_t\!\in\!\mathbb{R}^{d_a}\), without rewards or task
labels.
Figure~\ref{fig:qantara-modes} sketches the predictor and
Figure~\ref{fig:tau-square} the
sampling design (right).

\subsection{Learning the world-action model}
\label{sec:method-learning}

\paragraph{Model architecture.}
\label{sec:method-arch}
Qantara consists of an encoder \(E_\phi\) that maps each frame
\(o_t\) to a \(d\)-dimensional latent
\(z_t\!=\!E_\phi(o_t)\), and a predictor \(f_\theta\) that models
the joint distribution \(p_\theta\) of actions and future latents
along a trajectory. We organise the trajectory into a sequence of
\emph{blocks} indexed by \(t\!\in\!\{0, \ldots, T\}\), where each
block bundles latent \(z_t\) with the action that produced it; for
block \(0\) a learnable start-of-sequence token stands in for the
non-existent prior action. The predictor factors \(p_\theta\)
autoregressively across blocks but jointly within each block: at
step \(t\), the pair \((a_t,\, z_{t+1})\) of action and resulting
latent (which together populate block \(t\!+\!1\)) is predicted
jointly given the prefix \((z_{\leq t},\, a_{<t})\). For any clean
(un-noised) prefix length \(t'\!\in\!\{0,\ldots,T-1\}\), the chain
rule then gives
\begin{equation}
p_\theta(a_{t':T-1},\, z_{t'+1:T} \mid z_{\leq t'},\, a_{<t'}) \;=\; \prod_{t=t'}^{T-1} p_\theta(a_t,\, z_{t+1} \mid z_{\leq t},\, a_{<t}),
\label{eq:joint}
\end{equation}
and \(f_\theta\) parameterises each per-block conditional with a
transformer whose attention is block-causal across blocks
(realising Eq.~\ref{eq:joint}) and bidirectional within a block, so
the action and state tokens of a block are denoised jointly under
the bridge-flow objective detailed below.
\paragraph{Joint bridge-flow matching.}
\label{sec:method-bridge}
At training time the predictor reads a sequence split into a random observed context and a noised continuation it must denoise:
\begin{equation}
\big[\,
\underbrace{a_{-1}\, z_0\quad a_0\, z_1\quad \cdots\quad a_{t-1}\, z_{t}}_{\text{context (observed)}}\quad
\underbrace{\tilde a_{t}\, \tilde z_{t+1}\quad \cdots\quad \tilde a_{T-1}\, \tilde z_T}_{\text{noised (to denoise)}}
\,\big].
\label{eq:sequence}
\end{equation}
Each action and state token carries its own scalar
\(\tau\!\in\![0,1]\), indexed by the token's time
(\(\tau^a_t\) for action token \(a_t\), \(\tau^z_t\) for state
token \(z_t\)); \(\tau\!=\!1\) is the clean endpoint and
\(\tau\!=\!0\) the source endpoint. Each \(\tau\) is fed to the
predictor as a sinusoidal time-encoding through per-token AdaLN
modulation \citep{peebles2023scalable,perez2018film}. The two axes
use different interpolants because inference exposes a different
anchor at \(\tau\!=\!0\) on each. Actions have no anchor: at every
rollout step the planner queries the predictor for an action it
does not yet possess, so the \(\tau^a_t\!=\!0\) source must be
task-agnostic and we adopt the standard noise-to-data
flow-matching interpolant
\citep{lipman2022flow,liu2022flow,albergo2023stochastic,gao2025vita},
\begin{equation}
\tilde a_t \;=\; \tau^a_t\, a_t \;+\; (1-\tau^a_t)\,
\varepsilon_a, \qquad \varepsilon_a \sim \mathcal N(0, I).
\label{eq:action-fm}
\end{equation}
The latent, by contrast, always has the previous clean \(z_t\) as
a free anchor; pairing it with the unknown target \(z_{t+1}\) as
endpoints of a Brownian-bridge interpolant
\citep{albergo2023stochastic,liu2023ipsb,lyu2024framebridge} gives
\begin{equation}
\tilde z_{t+1} \;=\; (1-\tau^z_{t+1})\, z_t \;+\; \tau^z_{t+1}\, z_{t+1} \;+\;
\gamma\,\sqrt{\tau^z_{t+1} (1-\tau^z_{t+1})}\,\varepsilon_z, \qquad
\varepsilon_z \sim \mathcal N(0, I),
\label{eq:bridge}
\end{equation}
whose \(\tau^z_{t+1}\!=\!0\) source is exactly the previous latent
the planner feeds in at deployment. Because this source already
matches the deployment input, the model does not have to learn a
map from arbitrary Gaussian noise back to the latent manifold; it
only has to model the on-manifold transition \(z_{t+1}\!-\!z_t\)
between consecutive rollout steps.

\paragraph{Training objective.}
\label{sec:method-objective}
For each noisy target block, the transformer produces a hidden
vector per token, contextualised over earlier blocks (block-causal
attention) and the same-block partner (within-block bidirectional
attention):
\begin{equation}
(h^a_t,\, h^z_{t+1}) \;=\; f_\theta\big(\tilde a_{\le t},\, \tilde z_{\le t+1}\big).
\label{eq:hidden}
\end{equation}
The state head reads \(h^z_{t+1}\) and emits a residual on the
previous clean latent; the action head reads \(h^a_t\) and emits
an action velocity,
\begin{equation}
\hat z_{t+1} \;=\; z_t + \mathrm{head}_z\big(h^z_{t+1}\big),
\qquad
v^a_t \;=\; \mathrm{head}_a\big(h^a_t\big).
\label{eq:head-z}
\end{equation}
The state-head residual form (Eq.~\ref{eq:head-z}) pairs with
zero-initialised final-layer weights, making the predictor the identity
at initialisation (\(\hat z_{t+1}\!=\!z_t\)); we ablate the residual choice
in \S\ref{sec:exp-zdelta}. The state-axis bridge loss regresses the
clean endpoint and the action-axis flow-matching loss the velocity,
\begin{equation}
\mathcal{L}_{\mathrm{B}}^{\,z}
  = \mathbb{E}\,\|\hat z_{t+1}-z_{t+1}\|_2^2,
\qquad
\mathcal{L}_{\mathrm{FM}}^{\,a}
  = \mathbb{E}\,\|v^a_t-(a_t-\varepsilon_a)\|_2^2,
  \label{eq:loss-z}
\end{equation}
where the expectation is over data transitions
\((z_t, a_t, z_{t+1})\), flow times \((\tau^a_t, \tau^z_{t+1})\)
sampled per \S\ref{sec:method-modes}, and independent Gaussian
noises \(\varepsilon_a, \varepsilon_z\!\sim\!\mathcal N(0, I)\).
Optimising \(\mathcal{L}_{\mathrm{B}}^{\,z}\) alone admits a
collapse: if the encoder outputs the same latent for every frame,
the bridge target is trivially matched and the action head sees no
state information to condition on. We add SIGReg
\citep{balestriero2025lejepa}, the Sketched-Isotropic-Gaussian
regulariser, applied to the batch of clean latents
\(\{z_0, \ldots, z_T\}\) to drive their distribution toward an
isotropic Gaussian; the total objective is
\begin{equation}
\mathcal{L} \;=\; \lambda_z\,\mathcal{L}_{\mathrm{B}}^{\,z}
            \;+\; \lambda_a\,\mathcal{L}_{\mathrm{FM}}^{\,a}
            \;+\; \lambda_{\mathrm{SIG}}\,\mathrm{SIGReg}\big(\{z_0, \ldots, z_T\}\big),
\label{eq:loss-total}
\end{equation}
with \(\lambda_z\!=\!3\), \(\lambda_a\!=\!1\), and
\(\lambda_{\mathrm{SIG}}\!=\!0.09\). All components (encoder,
predictor, and the action and state heads) are trained jointly from
scratch. We adopt LeWM's no-heuristics training recipe
\citep{maes2025lewm}: no stop-gradient on encoder outputs, no EMA
target encoder, and no pretrained representations, with SIGReg as
the sole anti-collapse mechanism.

\subsection{Inference: three paradigms from a single checkpoint}
\label{sec:method-inference}

Every control step has the same setup: the encoder produces a
clean latent that extends the context prefix
\((z_{\le t},\, a_{<t})\), and we need to produce the next action
\(a_t\) to execute. Because \(a_t\) lives in the joint target block
\((a_t, z_{t+1})\), the same checkpoint admits three qualitatively
different ways to obtain it, depending on whether an encoded goal
vector is supplied at inference.

Geometrically, each query fixes one of
\((\tau^a_t, \tau^z_{t+1})\) at \(0\) or \(1\) and sweeps the
other from \(0\) to \(1\), tracing a single side of the
\([0,1]^2\) noise-time square (Figure~\ref{fig:tau-square}). We
call these four sides the \emph{edges} of the square and refer to
each dispatch by the edge it traverses (\texttt{policy},
\texttt{forward}, \texttt{video}, or \texttt{inverse}); training
concentrates capacity along them (\S\ref{sec:method-modes}).

\paragraph{Behaviour-cloning sampling.}
We sample \(a_t\) directly from the BC conditional
\(p_\theta(a_t\!\mid\!z_{0:t},\, \tilde z_{t+1}\!=\!z_t)\) along
the \texttt{policy} edge (\(\tau^z\!=\!0\),
\(\tau^a\colon 0\!\to\!1\)), where the predictor acts as a
state-conditional behaviour policy. With the state slot of block
\(t\!+\!1\) held at \(z_t\), the action slot starts as Gaussian
noise and \(K\) Euler steps on the velocity head sweep it to a
clean sample,
\begin{equation}
\tilde a^{(k+1)} \;=\; \tilde a^{(k)} \;+\; \tfrac{1}{K}\,
                  v^a\big(\tilde a^{(k)},\, \tau^a_t\!=\!k/K\big),
\qquad \tilde a^{(0)}\sim\mathcal N(0,I),\quad k = 0,\ldots,K\!-\!1,
\label{eq:bc-rollout}
\end{equation}
giving \(a_t\!=\!\tilde a^{(K)}\). At deployment we run
\(K\!=\!2\) Euler steps over a context window of length
\(T\!=\!4\) matching training.

\paragraph{Video--inverse composition.}
The BC conditional \(p_\theta(a_t\!\mid\!z_{0:t})\) also factorises
through the next latent, as in UniPi's video-as-policy paradigm
\citep{du2023unipi}; here the ``video'' factor is the predicted JEPA
latent \(\hat z_{t+1}\) (not a pixel frame), and no decoder runs at
inference. By the chain rule,
\begin{equation}
\underbrace{p_\theta(a_t\!\mid\!z_{0:t})}_{\text{BC dispatch}}
\;=\;
\int \underbrace{p_\theta(z_{t+1}\!\mid\!z_{0:t})}_{\texttt{video}\text{ edge}}
     \underbrace{p_\theta(a_t\!\mid\!z_{0:t},\, z_{t+1})}_{\texttt{inverse}\text{ edge}}
     \, \mathrm{d}z_{t+1},
\label{eq:video-inverse}
\end{equation}
which we approximate by point-evaluating at the conditional mean
\(\hat z_{t+1}\) of the video factor. The two factors traverse two
edges of the \((\tau^a, \tau^z)\) square. First, the
\texttt{video} edge (\(\tau^a\!=\!0, \tau^z\colon 0\!\to\!1\)):
with the action slot filled by pure Gaussian noise (carrying no
information about \(a_t\)), the predictor reduces to an action-free
latent dynamics model and reads off
\(\hat z_{t+1}\!\approx\!\mathbb E[z_{t+1}\!\mid\!z_{0:t},\,
\tilde z_{t+1}\!=\!z_t]\), the action-marginal next latent under
the training-time action prior. We follow the JEPA world model convention
in calling this the \texttt{video} edge
\citep{zhou2024dinowm,assran2025vjepa2,maes2025lewm}; we operate
in latent space and never decode pixels. Second, the
\texttt{inverse} edge (\(\tau^z\!=\!1, \tau^a\colon 0\!\to\!1\)):
with both transition endpoints clean (\(z_t\) on the prefix and
\(\hat z_{t+1}\) on the state slot, action slot from Gaussian
noise), \(K\) Euler steps recover
\(a_t\!\sim\!p_\theta(a_t\!\mid\!z_{0:t},\,
\tilde z_{t+1}\!=\!\hat z_{t+1})\), the inverse-dynamics action
that bridges \(z_t\) to \(\hat z_{t+1}\). The composition gains
\(1\)--\(2\)\,SR over BC sampling on Push-T and Cube, and ties
within \(\pm 1.5\)\,SR on Two-Room and Reacher
(\S\ref{sec:exp-bc}).

\paragraph{Latent planning.}
Given a goal observation \(o_g\) encoded to
\(z_g\!=\!E_\phi(o_g)\), we search for an action sequence
\(a^*_{0:H-1}\) minimising the terminal cost
\(\mathcal C(\hat z_H)\!=\!\|\hat z_H\!-\!z_g\|_2^2\), following
LeWM~\citep{maes2025lewm}. Each rollout step queries the
\texttt{forward} edge (\(\tau^a\!=\!1, \tau^z\colon 0\!\to\!1\)):
with the candidate clean action \(a_t\) on the action slot and
the previous latent \(z_t\) on the state slot, the predictor acts
as a forward dynamics model and reads off
\(\hat z_{t+1}\!\approx\!\mathbb E[z_{t+1}\!\mid\!z_{0:t},\, a_t,\,
\tilde z_{t+1}\!=\!z_t]\) from the residual head
(Eq.~\ref{eq:head-z}). We iterate the predictor recursion for
\(K\) denoising steps along \(\tau^z\colon 0\!\to\!1\) to
sharpen the \(\hat z_{t+1}\) estimate (\(K\!=\!4\); flat sweep
over \(\{1,2,4,8\}\) in App.~\ref{app:k-sweep-cem}). The search uses the
Cross-Entropy Method~\citep{rubinstein2004cross} with \(N\)
Gaussian-proposal candidates and elite-set refit; rollouts
accumulate prediction error with horizon, so we apply
receding-horizon Model Predictive Control, executing the
elite-mean plan before replanning
(App.~\ref{app:cem-mpc-config}).

\subsection{Edge-aligned \texorpdfstring{$(\tau^a,\tau^z)$}{(tau\_a, tau\_z)} sampling}
\label{sec:method-modes}

\begin{wrapfigure}[21]{r}{0.46\linewidth}
  \centering
  \includegraphics[width=0.98\linewidth,trim=0 22 0 0,clip]{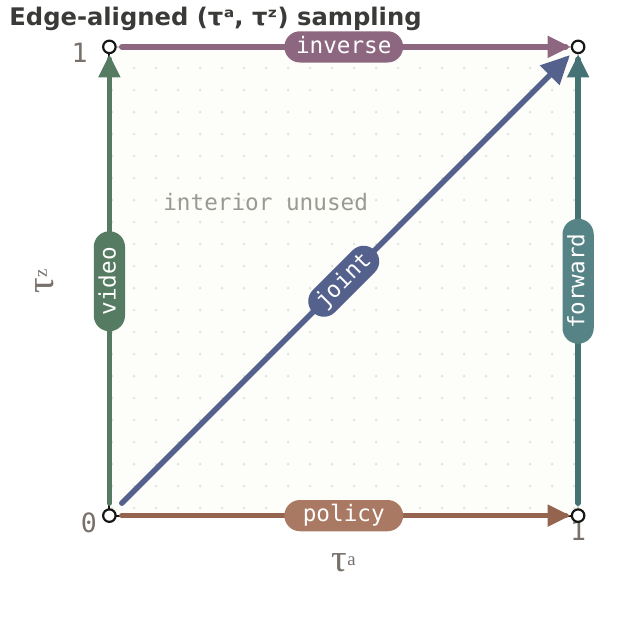}
  \caption{\textbf{Edge-aligned \((\tau^a, \tau^z)\) sampling.}
  Per-token \((\tau^a, \tau^z)\) are drawn under
  five 1D modes: the four edges of \([0,1]^2\) (\texttt{forward},
  \texttt{policy}, \texttt{inverse}, \texttt{video}) and the
  \texttt{joint} diagonal; the 2D interior is never sampled,
  so capacity stays on the regions queried at inference.}
  \label{fig:tau-square}
\end{wrapfigure}

Sampling \((\tau^a_t, \tau^z_{t+1})\) uniformly on \([0,1]^2\)
during training would spend most capacity on the interior, where
no inference dispatch ever queries. We concentrate training on the
four edges instead, plus a fifth \texttt{joint} mode along the
diagonal (\(\tau^a\!=\!\tau^z\), sweeping \(0\!\to\!1\) together).
The joint diagonal is never queried at inference, but removing it
destabilises training (\S\ref{sec:exp-modes}).

Each batch row is replicated across all five modes
(Figure~\ref{fig:tau-square}); within each mode the swept axis
follows a per-block monotone-decreasing \(\tau\)-chain
\citep{xie2025pavdm,liu2025rollingforcing} and a random clean
prefix keeps every cold-start rollout horizon in-distribution
\citep{po2025colds,zhu2026causalforcing}, both detailed in
App.~\ref{app:tauchain}.

\section{Experiments}
\label{sec:experiments}

\subsection{Setup}
\label{sec:exp-setup}

We evaluate
on the \emph{LeWM-suite}~\citep{maes2025lewm}, the four-environment
subset on which LeWM places PLDM, DINO-WM, and itself in a single
end-to-end pixel-JEPA pipeline.
The subset supplies apples-to-apples baseline numbers, and its four
envs cross 2D vs.\ 3D observations, navigation vs.\ manipulation
vs.\ motor control, and four classes of data-collection oracle
(noisy heuristic, human demonstration, scripted expert, uniform
random), so no single data distribution dominates the suite
average. The only broader pixel JEPA-world-model benchmark,
DINO-WM's six-environment evaluation~\citep{zhou2024dinowm}, adds
envs either in categories this suite already covers or in a
deformable-dynamics class separate from the multi-paradigm
question.
The four envs are \textbf{Two-Room}~\citep{sobal2025stresstesting}
(2D navigation), \textbf{Push-T}~\citep{chi2023diffusion} on the
DINO-WM-released expert dataset~\citep{zhou2024dinowm} (contact-rich
2D manipulation), \textbf{OGBench-Cube}~\citep{park2025ogbench} (3D
pick-and-place; OGBench's single-cube \texttt{cube-single} task,
App.~\ref{app:envs}), and \textbf{Reacher-Hard} from the DeepMind Control
suite~\citep{tassa2018deepmind} (precision motor control); per-env
trajectory counts, lengths, and oracles are in App.~\ref{app:envs}.
The three inference modes place different demands on data quality:
BC sampling cannot exceed the demonstrating policy and
video--inverse composition needs near-expert coverage, while CEM
planning needs only goal-spanning data. The near-expert corpora
meet all three on Two-Room, Push-T and Cube; Reacher's released
corpus is random instead (App.~\ref{app:envs}), and it behaves as
this predicts, planning still reaching 80.9\,SR while the two
imitation paths fall to or below the random-policy floor
(\S\ref{sec:exp-bc}).

Models are trained for 10 epochs per env. Each (recipe, env) cell
aggregates 50 episodes \(\times\) 5 eval seeds \(\times\) 3 train
seeds = 750 episodes; we report mean and population std across the
three train-seed means, matching \citet{maes2025lewm}'s 3-train-seed
convention with 5\(\times\) more eval episodes per train seed.

\subsection{Goal-conditioned planning on the LeWM suite}
\label{sec:exp-main}

We compare against the three end-to-end pixel-based world-model
baselines reported on this suite by \citet{maes2025lewm}:
PLDM~\citep{sobal2025stresstesting},
DINO-WM~\citep{zhou2024dinowm} (no-proprio variant: pixel-only
inputs, matching the proprio-free setting of the other
baselines), and LeWM~\citep{maes2025lewm} itself. Our training
and evaluation pipeline extends LeWM's code, so we also
retrain both LeWM and PLDM end-to-end in this pipeline.
DINO-WM numbers are taken from~\citet{maes2025lewm}.

\begin{table*}[t]
\small
\centering
\caption{\textbf{LeWM-suite results.} Per-environment success
rate; mean \(\pm\) population std across 3 train seeds.
\emph{Bold = best within our reproduction pipeline} (rows below
the midrule); the top three rows are numbers reported
by~\citet{maes2025lewm}, run under their pipeline, which trains
every model for 10 epochs and evaluates all methods under a fixed 50-step episode
budget, stricter than the original DINO-WM evaluation
(DINO-WM reports 90.0 on Push-T under its own protocol).
We retrain both LeWM and PLDM end-to-end in our pipeline, so the
bottom block shares data, training scaffold and eval budget across
all three rows. Both reproductions trail their published Reacher
numbers by almost the same margin (\(-21.7\) and \(-22.9\,\)SR)
while landing within \(5.6\,\)SR elsewhere, locating that gap
outside either method (see below). PLDM's loss recipe (one of its
two published ones) is chosen per environment on a single train seed
(App.~\ref{app:optim}).}
\label{tab:lewm-main}
\begin{tabular}{lcccc}
\toprule
method & Push-T & Two-Room & Cube & Reacher \\
\midrule
PLDM \citep{sobal2025stresstesting}     & 78.0\,\(\pm\)\,5.0 & 97.0\,\(\pm\)\,1.2 & 65.0\,\(\pm\)\,2.8 & 78.0\,\(\pm\)\,5.4 \\
DINO-WM \citep{zhou2024dinowm}          & 74.0\,\(\pm\)\,4.5 & 100.0\,\(\pm\)\,0.0 & 86.0\,\(\pm\)\,4.7 & 79.0\,\(\pm\)\,5.1 \\
LeWM \citep{maes2025lewm} (paper)       & 96.0\,\(\pm\)\,4.0 & 87.0\,\(\pm\)\,2.5 & 74.0\,\(\pm\)\,3.0 & 86.0\,\(\pm\)\,5.0 \\
\midrule
PLDM (reproduction)                     & 72.4\,\(\pm\)\,1.8 & 96.9\,\(\pm\)\,1.0 & 64.5\,\(\pm\)\,1.9 & 55.1\,\(\pm\)\,1.5 \\
LeWM (reproduction)                     & \textbf{92.3}\,\(\pm\)\,2.0 & 89.9\,\(\pm\)\,1.4 & 76.7\,\(\pm\)\,1.3 & 64.3\,\(\pm\)\,1.7 \\
\textbf{Qantara (ours)}                 & 90.1\,\(\pm\)\,1.1 & \textbf{100.0}\,\(\pm\)\,0.0 & \textbf{93.7}\,\(\pm\)\,0.7 & \textbf{80.9}\,\(\pm\)\,1.8 \\
\bottomrule
\end{tabular}
\end{table*}

Per-environment, Qantara ties the Two-Room ceiling with DINO-WM
and exceeds the DINO-WM row on Push-T (90.1 vs 74.0 re-evaluated)
and Reacher (80.9 vs 79.0), trailing LeWM-paper by 5.9 and 5.1 on
those two. An independent evaluation of LeWM's released
checkpoint under its default planner and the published 50-episode
size scores \(89.3\,\pm\,6.4\) on Push-T and \(68.0\,\pm\,9.2\) on
Reacher~\citep{nguyen2026gcidm}, close to our LeWM reproduction
(92.3 and 64.3). Retraining PLDM here shows
the same pattern: within 5.6\,SR of its published Push-T, Two-Room
and Cube numbers but \(-22.9\) on Reacher.

Two measurements place the Reacher gap in the training budget:
a random policy scores 11.2 under our protocol against the published
10.9~\citep{maes2025lewm}, and at 20 epochs the LeWM reproduction
rises from 64.3 to \(80.0\,\pm\,0.9\) while Qantara stays flat
(\(79.7\,\pm\,1.8\)); PLDM recovers to \(78.4\,\pm\,3.3\)
(App.~\ref{app:reacher-budget}). The Reacher column is therefore a
matched-10-epoch statement, and the \(+16.6\) Reacher margin below
does not persist at the longer budget.

Head-to-head against the LeWM reproduction at this budget, Qantara
wins on three of four envs (Two-Room \(+10.1\), Cube \(+17.0\),
Reacher \(+16.6\), the last of which the 20-epoch control above
erases) and trails by 2.2\,SR on Push-T. The Cube margin survives
two controls trained in the same pipeline over three train seeds
(App.~\ref{app:optim}): DINO-WM retrained here reaches
\(80.0\,\pm\,1.5\) (\(+13.7\) for Qantara), and a LeWM predictor
grown to 20.75M parameters, against Qantara's 21.55M, reaches \(74.7\,\pm\,1.6\), below the stock
76.7. On the harder two-cube variant (Cube-Double,
App.~\ref{app:envs}) the planners fall to \(49.6\,\pm\,1.0\) for
Qantara against \(36.9\,\pm\,0.7\) for LeWM and \(36.3\,\pm\,0.5\)
for DINO-WM, all three trained in our pipeline.

\subsection{The same checkpoint serves three inference paradigms}
\label{sec:exp-bc}

Tab.~\ref{tab:bc-vs-cem} reports all three paths on Push-T,
Two-Room and Cube, queried from the same Qantara checkpoints as
Tab.~\ref{tab:lewm-main} at the inference defaults of
\S\ref{sec:method-inference} (CEM \(K\!=\!4\); BC and
video--inverse \(K\!=\!2\)) without per-paradigm retraining.

\begin{table}[ht]
\small
\centering
\caption{\textbf{One Qantara checkpoint, three inference paradigms.}
Mean SR \(\pm\) std across 3 train seeds. CEM is goal-conditioned
(it plans against the encoded goal \(z_g\)); BC and video--inverse
are goal-blind imitation paths that receive no goal input and run
at 17--27\,ms per env-step versus CEM's 0.4--1.2\,s. Reacher
omitted: the goal-blind dispatches drop to near-noise there
(6.1 / 4.8\,SR for BC / video--inverse, see the text); its CEM
result is in Tab.~\ref{tab:lewm-main}.
The GCBC column (goal-conditioned behaviour cloning,
\citealp{ghosh2019learning}) is the suite authors' own number, run
on DINOv2 patch embeddings rather than retrained in our pipeline,
so it is a cross-pipeline reference and not a matched comparison.
}
\label{tab:bc-vs-cem}
\resizebox{\textwidth}{!}{%
\begin{tabular}{lcccc}
\toprule
            & \multicolumn{3}{c}{Qantara (single checkpoint)} & \multirow{2}{*}{GCBC} \\
\cmidrule(lr){2-4}
environment & CEM (\(K\!=\!4\)) & BC (\(K\!=\!2\), goal-blind) & video--inverse (\(K\!=\!2\), goal-blind) & \\
\midrule
Push-T   & 90.1\,\(\pm\)\,1.1  & 82.1\,\(\pm\)\,1.0 & 83.2\,\(\pm\)\,0.6 & 75   \\
Two-Room & 100.0\,\(\pm\)\,0.0 & 69.1\,\(\pm\)\,0.4 & 69.3\,\(\pm\)\,0.8 & 100  \\
Cube     & 93.7\,\(\pm\)\,0.7  & 70.8\,\(\pm\)\,0.3 & 72.8\,\(\pm\)\,0.6 & 84   \\
\bottomrule
\end{tabular}%
}
\end{table}

\textbf{The goal-blind dispatches succeed when the demonstration
distribution supplies the missing goal information and the env's
success tolerance absorbs the imitation drift.} In every env the
eval goal is a later state of the dataset trajectory an episode starts
from (App.~\ref{app:envs}). On Push-T that trajectory heads for the task
goal fixed across the corpus, which a goal-blind sampler internalises
in its weights rather than reading it off pixels (masking
the drawn marker at evaluation changes BC and video--inverse by
at most 2\,SR, within seed noise). BC and video--inverse reach
82.1 and 83.2\,SR, above the 75 the suite authors report for
goal-aware GCBC (see the caption), and match the 1-mode \texttt{policy}
specialist trained in the same pipeline (81.3, Tab.~\ref{tab:mode-set}); CEM uses
\(z_g\) to recover the full 90.1\,SR. On
Two-Room and Cube the per-episode goal is randomised and never
rendered, so BC and video--inverse reach it by reproducing the
expert's continuation. On
Reacher the per-episode target qpos requires precise joint
matching that the goal-blind dispatches cannot reliably deliver,
so BC and video--inverse drop to near-noise (6.1 / 4.8\,SR;
omitted from Tab.~\ref{tab:bc-vs-cem}).
The goal-blind paths are therefore the fast dispatch wherever the
task goal is fixed across demonstrations (Push-T) or the success
tolerance absorbs imitation drift (Two-Room, Cube); on precision
targets such as Reacher, planning is the reliable dispatch. On Cube-Double the ordering flips: the
goal-blind paths reach \(54.3\,\pm\,1.4\) and \(54.4\,\pm\,2.3\),
above the planner's 49.6, so the best dispatch depends on the task.

At the inference defaults BC and video--inverse run
\(\sim\!15\)--\(65\!\times\) faster than CEM
(App.~\ref{app:timing});
decoded rollouts under all three dispatches stay coherent from the
same checkpoint (App.~\ref{app:rollouts}).

\textbf{A joint video--action model under the same protocol.}
We train UVA \citep{li2025uva} on our Push-T corpus at a matched
optimiser-step budget (248{,}268--248{,}336 steps against the
released recipe's 247{,}050) and at its published inference
cadence, then query its action head through the same evaluation
code, episode budget and five evaluation seeds as every row
above. Over three training seeds it reaches
\(14.4\,\pm\,1.8\,\)SR, against \(82.1\) for our goal-blind BC
path and \(1.6\,\pm\,2.0\) for a random policy. The UVA
checkpoint fits its training data: run open-loop on frames from its
own training corpus, it imitates the expert's next action chunk to
\(0.0156\) mean absolute error in a two-unit-wide action space
(\(18\)--\(21\!\times\) better than reversed- and shuffled-frame
controls), so the gap arises in closed-loop control. On one UVA
training seed, replanning at every step instead of at its published
cadence recovers only \(+5.6\,\)SR (paired \(95\%\) CI \([+1.1,+10.1]\)), about \(8\%\)
of the gap.

\subsection{Which design choices matter}
\label{sec:exp-ablations}

\paragraph{Edge-aligned sampling beats uniform coverage.}
\label{sec:exp-uniform}
Uniform sampling on \([0,1]^2\), the independent per-modality
noise levels UWM trains with \citep{zhu2025uwm}, places measure
zero on the four edges that inference queries
(\S\ref{sec:method-inference}), and the cost falls hardest on the
action-sensitive cells.
Replacing the five edge-aligned modes
(\S\ref{sec:method-modes}) with five copies of a uniform sampler
at matched compute, with the bridge and all else fixed, regresses Push-T CEM by \textbf{36.8\,SR}
(90.1 \(\to\) 53.3), Push-T video--inverse by 33.5\,SR
(83.2 \(\to\) 49.7), and Reacher CEM by 23.4\,SR
(80.9 \(\to\) 57.5); Cube CEM regresses modestly by 6.8\,SR
(93.7 \(\to\) 86.9) and Push-T BC stays within seed noise
(82.1 \(\to\) 81.6) (Table~\ref{tab:mode-set}, second row).

\paragraph{Mode-set ablation: trunk co-regularisation across modes.}
\label{sec:exp-modes}

At a fixed batch of examples and optimiser-step count, we ablate
the 5-mode design (\S\ref{sec:method-modes}) and evaluate all
three inference paths per cell (Tab.~\ref{tab:mode-set}; 11-recipe sweep). Per-mode
coverage predicts that dropping a mode breaks only the inference
path querying its corner; three findings on Push-T overturn this.

\begin{table}[ht]
\footnotesize
\centering
\setlength{\tabcolsep}{3pt}
\caption{\textbf{Mode-set + sampler ablation at the default recipe},
multi-seed (3 train \(\times\) 5 eval \(\times\) 50 episodes).
Inference defaults from \S\ref{sec:method-inference}: CEM
\(K\!=\!4\), BC and video--inverse \(K\!=\!2\).
\(\dagger\): train-seed collapse (pop-std \(>\,15\,\)SR). Two-Room
omitted (CEM reaches \(100\,\)SR at the reference,
Tab.~\ref{tab:bc-vs-cem}; no headroom for ablation).
BC and video--inverse omit Cube and Reacher, where both goal-blind
paths are insensitive to the mode set (Cube, 66--73\,SR in every
evaluated row but 1-mode \{\texttt{forward}\}) or collapse to
near-noise (Reacher). The 1- and
2-mode rows double as the specialist checkpoints, each trained on
that locus alone at the same recipe and step count. A row with \(M\)
modes trains on an effective batch of \(M\!\cdot\!B\), so rows with
equal \(M\) are compute-matched.
}
\label{tab:mode-set}
\qtfit{%
\begin{tabular}{l *{5}{r@{\(\,\pm\,\)}l}}
\toprule
& \multicolumn{6}{c}{CEM} & \multicolumn{2}{c}{BC} & \multicolumn{2}{c}{video--inverse} \\
\cmidrule(lr){2-7}\cmidrule(lr){8-9}\cmidrule(lr){10-11}
modes (sampler) & \multicolumn{2}{c}{Push-T} & \multicolumn{2}{c}{Cube} & \multicolumn{2}{c}{Reacher} & \multicolumn{2}{c}{Push-T} & \multicolumn{2}{c}{Push-T} \\
\midrule
5-mode (reference)                           & \textbf{90.1} & 1.1                  & 93.7          & 0.7 & \textbf{80.9} & 1.8                  & 82.1          & 1.0                  & \textbf{83.2} & 0.6 \\
5-mode uniform                               & 53.3          & 9.6                  & 86.9          & 1.0 & 57.5          & 6.9                  & 81.6          & 1.0                  & 49.7          & 1.3 \\
4-mode drop-\{\texttt{policy}\}              & 27.2          & 32.5\(^{\dagger}\)   & 89.9          & 4.5 & 82.3          & 4.0                  & 20.4          & 21.3\(^{\dagger}\)   & 21.7          & 22.9\(^{\dagger}\) \\
4-mode drop-\{\texttt{inverse}\}             & 69.1          & 23.7\(^{\dagger}\)   & 89.9          & 0.9 & 78.9          & 2.2                  & 65.5          & 25.6\(^{\dagger}\)   & 35.3          & 15.7\(^{\dagger}\) \\
4-mode drop-\{\texttt{video}\}               & 87.3          & 1.7                  & 93.7          & 0.2 & 80.1          & 1.7                  & 81.6          & 1.2                  & 81.6          & 0.3 \\
4-mode drop-\{\texttt{joint}\}               & 61.7          & 41.1\(^{\dagger}\)   & 91.2          & 2.3 & 76.7          & 2.2                  & 56.7          & 38.1\(^{\dagger}\)   & 56.8          & 37.3\(^{\dagger}\) \\
3-mode \{\texttt{forward, policy, inverse}\} & 76.0          & 21.8\(^{\dagger}\)   & \textbf{93.9} & 1.0 & 78.5          & 0.2                  & 68.4          & 19.5\(^{\dagger}\)   & 65.5          & 20.3\(^{\dagger}\) \\
2-mode \{\texttt{forward, policy}\}          & 88.5          & 1.4                  & 91.9          & 0.2 & 76.4          & 0.6                  & \textbf{82.3} & 0.8                  & 7.5           & 1.5 \\
2-mode \{\texttt{video, inverse}\}           & 10.4          & 7.7                  & 67.9          & 4.8 & 26.3          & 2.2                  & 20.5          & 9.2                  & 37.6          & 22.4\(^{\dagger}\) \\
1-mode \{\texttt{forward}\}                  & 62.0          & 33.9\(^{\dagger}\)   & 76.3          & 7.6 & 56.0          & 23.4\(^{\dagger}\)   & 2.4           & 0.0                  & 2.0           & 0.0 \\
1-mode \{\texttt{policy}\}                   & 2.1           & 0.7                  & 50.3          & 1.5 & 10.4          & 1.0                  & 81.3          & 1.3                  & 4.4           & 1.0 \\
1-mode uniform                               & 40.5          & 23.3\(^{\dagger}\)   & 84.9          & 1.5 & 51.2          & 7.4                  & 48.5          & 27.5\(^{\dagger}\)   & 27.9          & 15.8\(^{\dagger}\) \\
\bottomrule
\end{tabular}%
}
\end{table}

Of the four compute-matched drop-one rows, only \texttt{drop-video}
preserves all three Push-T paths, staying
within \(2.8\,\)SR of the 5-mode reference: the \((0,0)\) corner
where video--inverse starts denoising is still trained as the
lower endpoint of the \texttt{joint} diagonal and of the
\texttt{policy} edge, so the third path
survives without a dedicated mode. Dropping \texttt{policy},
\texttt{inverse}, or \texttt{joint} instead destabilises training
across all three paths, CEM pop-std exceeding \(15\,\)SR with
1--2 of three seeds collapsing below \(40\,\)SR (sharpest:
\texttt{drop-policy} \(27.2\!\pm\!32.5\) vs \texttt{drop-video}
\(87.3\!\pm\!1.7\)); the \((1,0)\) corner is the only point where
\texttt{policy}'s \(\tau^z\!=\!0\) edge and CEM's
\(\tau^a\!=\!1\) query meet, yet removing \texttt{policy} still
destabilises CEM, while Cube and Reacher hold within
\(\sim\!5\,\)SR of the reference across the four drops.
Neither tested 2-mode subset is self-sufficient:
\{\texttt{forward}, \texttt{policy}\} collapses on video--inverse
(\(7.5\,\)SR) and \{\texttt{video}, \texttt{inverse}\}
underperforms that same target (37.6 vs 83.2).

All five modes are therefore kept for \emph{Push-T training
stability}; \texttt{drop-video} already preserves the third path.
On the collapsed seeds,
final \(\mathcal{L}_{\mathrm{B}}^{\,z}\) ends 2--40\(\times\) above the
reference and separates collapsed from healthy drop-one cells without
overlap, so the instability is a training-time failure, not an
eval-only artefact.

\paragraph{Output residual + zero-init is load-bearing.}
\label{sec:exp-zdelta}
Ablating the state-head residual at fixed bridge interpolant, so
the predictor is no longer the identity at initialisation
(Eq.~\ref{eq:head-z}), regresses suite average by \(5.6\,\)SR and
Push-T by \(15.2\,\)SR, with seed-level training instability
(App.~\ref{app:z-delta}).

\paragraph{Defaults robust to \(K\); off-canonical \(\gamma\) collapses.}
\label{sec:exp-robustness}
Sweeping the predictor recursion step count
\(K\!\in\!\{1,2,4,8\}\) leaves the suite average within seed noise
(89.8--91.2\,SR), and the \(\gamma\!=\!\sqrt 2\)
Schr\"odinger-bridge canonical \citep{liu2023ipsb} ties our
\(\gamma\!=\!1\) stochastic-interpolant canonical
\citep{albergo2023stochastic} at 90.5 vs.\ 90.8\,SR; off-canonical
\(\gamma\!\in\!\{0,0.5,2\}\) collapse 1--2 of 3 train seeds. We
retain \(\gamma\!=\!1\), and drop the upstream LeWM predictor
dropout (\(p\!=\!0.1\!\to\!0\); \(+3.4\,\)SR suite average at an earlier recipe). Full
sweeps: App.~\ref{app:robustness}.

\section{Related work}
\label{sec:related}

\paragraph{World models for control.}
JEPA-style world models predict environment dynamics in a learned embedding
space: PLDM \citep{sobal2025stresstesting} trains end-to-end with
VICReg, DINO-WM \citep{zhou2024dinowm} freezes DINOv2, LeWM
\citep{maes2025lewm,balestriero2025lejepa} pairs next-embedding
prediction with an isotropic-Gaussian regulariser, and V-JEPA-2-AC
\citep{assran2025vjepa2} post-trains a 300M action-conditional
predictor on a frozen 1B video JEPA for model-predictive control.
Recent extensions push the recipe to dexterous manipulation
\citep{goswami2025dexwm}, one-shot imitation \citep{goswami2025osviwm},
masked latent interventions \citep{nam2026cjepa}, value-aligned
latent shaping \citep{destrade2026vjepa}, and JEPA world model design-choice studies
\citep{terver2025drives}. On this suite \citet{gao2026fastlewm} cut
CEM solve time with action-prefix rollout and \citet{nguyen2026gcidm}
amortise the search into a head fitted on frozen LeWM latents,
reporting higher search-free success than ours on Cube and Reacher. Adjacent world models include TD-MPC2
\citep{hansen2024tdmpc2,hansen2025mmworld}, DreamerV4
\citep{hafner2025dreamerv4}, and NWM \citep{bar2025navigation};
TD-MPC2 pairs a planner with a policy prior, the rest serve one
paradigm. Qantara serves planning, behaviour cloning, and inverse
dynamics from one checkpoint under a single objective.

\paragraph{Joint state-action denoising at billion scale.}
A second cluster trains one transformer to denoise observations and
actions jointly and serves forward dynamics, inverse dynamics, and
behaviour policy from one checkpoint, predicting the future
observation rather than latent dynamics of their own:
PAD \citep{guo2024pad} (joint-denoising DiT), UVA
\citep{li2025uva} (decoupled video and action heads), UWM
\citep{zhu2025uwm} (per-modality timesteps), DUST \citep{won2025dust}
(dual-stream MMDiT), WorldVLA / RynnVLA-002
\citep{cen2025worldvla,cen2025rynnvla} (autoregressive token
transformer), Cosmos Policy \citep{kim2026cosmospolicy}
(post-trained 2B video model), and DreamZero
\citep{ye2026dreamzero}, a 14B autoregressive video-diffusion
backbone whose flow-matching objective shares one denoising
timestep across the video and action modalities. GR-1
\citep{wu2024gr1} is the
GPT-style ancestor. Parameter counts span 150M to 14B; all
reconstruct the observation except DUST, which generates frozen-VLM
embeddings of it, and none learns latent dynamics of its own.
Trajectory-diffusion and per-token-noise antecedents
\citep{janner2022diffuser,ajay2023dd,chen2024dforcing} are
reviewed in App.~\ref{app:diffusion-antecedents}.
Table~\ref{tab:va-capability} makes the capability structure of
this cluster explicit: goal-conditioned test-time planning is the
row no member of it exposes (UVA only re-ranks another policy's
proposals), because trajectory optimisation queries the predictor
hundreds of times per control step, which a compact latent space
makes cheap. We also train the cluster's
UVA on our Push-T corpus and evaluate it under our own protocol
(\S\ref{sec:exp-bc}).

\paragraph{Position.}
The \emph{(JEPA-latent, multi-paradigm)} cell of the
(representation \(\times\) paradigm-count) plane the two clusters
tile is sparse \citep{nguyen2026gcidm}. Qantara
reaches it from one joint objective and a single
\(\sim\!21\)M-parameter checkpoint, up to three orders of
magnitude smaller than the second cluster, trading raw capacity for
low-latency deployment (17--27\,ms per goal-blind step,
\S\ref{sec:exp-bc}).

\section{Conclusion}
\label{sec:conclusion}

Qantara trains a single \(\sim\!21\)M-parameter JEPA world model
and serves three inference paradigms from one set of weights. On the LeWM-suite the planning path
averages 91.2\,SR and lifts OGBench-Cube to 93.7\,SR, \(+7.7\)
over the suite's DINO-WM row
(Tab.~\ref{tab:lewm-main});
from the same checkpoint the behaviour-cloning and video--inverse
paths reach 82.1 and 83.2\,SR on Push-T at
\(\sim\!15\)--\(65\!\times\) lower inference cost than the planner
(\S\ref{sec:exp-bc}, Tab.~\ref{tab:bc-vs-cem}).

The design follows a single rule: training should match how the
model is queried at inference. The three dispatches read the
\((\tau^a, \tau^z)\) square only on its edges and corners, so
training samples there; the planner exposes the previous clean
latent at \(\tau^z\!=\!0\), so the state axis is a Brownian bridge
to that anchor. On the most action-sensitive environment, uniform
\([0,1]^2\) sampling at matched compute regresses Push-T CEM by
36.8\,SR (\S\ref{sec:exp-uniform}), and ablating the state-head
residual\,+\,zero-init pair, a separate choice, regresses it by
15.2\,SR (\S\ref{sec:exp-zdelta}); replacing the bridge interpolant
with a Gaussian source costs 12.7\,SR on Reacher with the residual
off (App.~\ref{app:z-delta}). The mode-set ablation then produced a
result we did not expect: dropping any single \texttt{policy},
\texttt{inverse}, or \texttt{joint} mode destabilises Push-T
training across \emph{all three} inference paths, including the
planner, whose edge meets the dropped mode's at a single corner at most
(\S\ref{sec:exp-modes}, Tab.~\ref{tab:mode-set}), so the five modes
co-regularise the shared trunk beyond their per-corner coverage
role.

\paragraph{Limitations.}
The capability claim is bounded on three axes: single-block
denoising per control step, with chunked multi-block inference
\citep{chen2024dforcing,po2025colds} the direct extension;
simulation-only evaluation on near-expert offline data, leaving
real-robot transfer and natural-video pre-training
\citep{assran2025vjepa2} open; and a single
\(\sim\!21\)M-parameter scale.

%% file: appendix.tex
\appendix

\section*{Appendix}
\addcontentsline{toc}{section}{Appendix}

\section{Environments and datasets}
\label{app:envs}

All four envs use continuous action spaces and are goal-conditioned
on a future state sampled from the same trajectory.
\textbf{Two-Room}~\citep{sobal2025stresstesting} is 2D continuous
navigation through a wall-with-door (\(10\)k trajectories, average
\(92\) steps, generated by a noisy door-then-target heuristic).
\textbf{Push-T}~\citep{chi2023diffusion} is contact-rich 2D
manipulation reorienting a T-block; we use the DINO-WM-released
dataset~\citep{zhou2024dinowm} of \(20\)k human-designed expert
demonstrations averaging \(196\) steps.
\textbf{OGBench-Cube}~\citep{park2025ogbench} refers to OGBench's
cube-single task (one cube, \(10\)k expert trajectories of \(200\)
steps); Cube in this paper always denotes this single-cube variant.
Trajectories are generated by the OGBench scripted data-collection
policy. \textbf{Cube-Double} is the two-cube variant of the same
environment; following the recipe behind the single-cube corpus, we
generate \(10\)k trajectories of \(200\) steps at \(224\)\,px
with the same scripted oracle and collection mode, and plan and
evaluate with the Cube settings unchanged.
\textbf{Reacher-Hard} from the DeepMind Control
suite~\citep{tassa2018deepmind} requires precise two-joint alignment,
\(10\)k trajectories of \(200\) steps. We use the Reacher corpus that
the LeWM codebase and \texttt{stable-worldmodel} ship, which is
collected by a uniform random policy rather than by the Soft
Actor-Critic policy the LeWM appendix describes. Measured on the
released file: the action marginal is uniform to within \(0.1\%\)
over ten bins, mean finger-to-target distance is flat across the
episode (\(0.203\) at step 0, \(0.204\) at step 200), and only
\(3.8\%\) of steps fall inside the success tolerance. Two
observations indicate the published Reacher numbers rest on this
same corpus rather than on a Soft Actor-Critic one: it is the only
Reacher dataset either shipped configuration names, and the
random-policy control agrees across the two pipelines
(\(11.2\,\pm\,4.8\) here against a published \(10.9\,\pm\,5.3\)),
which a goal distribution drawn from a converged reaching policy
would not reproduce.

\section{Optimisation and training-time hyperparameters}
\label{app:optim}

We train with AdamW \citep{loshchilov2019adamw}: weight decay \(10^{-3}\),
gradient clip \(1.0\), batch size \(128\), bf16 mixed precision, peak learning
rate \(3\!\times\!10^{-4}\) under a linear-warmup cosine-annealing schedule
(\(1\%\) warmup, cosine decay to \(0\) over the remaining steps). All five
training modes (\S\ref{sec:method-modes}) are enabled by default, so the
effective batch is \(5\!\cdot\!B\!=\!640\).

One environment trains in roughly 2--3 hours on a single H100 for
the \(\sim\!21\)M-parameter model (10 epochs,
\(\approx\!50\)k optimiser steps at \(B\!=\!128\)). The LeWM
baseline row in Tab.~\ref{tab:lewm-main} was retrained in this
same pipeline with the same data, \(B\!=\!128\) source rows per step
and optimiser-step count, as was the PLDM reproduction row
(16 training runs: two loss recipes on train seed 11 to select one
per environment, then the winner on seeds 22 and 33). The DINO-WM
row and the top-block PLDM row are as published
by \citet{maes2025lewm}. The two Cube controls of
\S\ref{sec:exp-main} are trained in this pipeline for 10 epochs
over three train seeds: DINO-WM with the recipe shipped in the
suite's codebase (frozen DINOv2-S/14 encoder, depth-6 predictor,
batch size \(32\), pixel-only inputs as in the suite's DINO-WM
row), and LeWM with its predictor grown from depth 6 and MLP width
2048 to depth 7 and width 2400 (20.75M parameters excluding the
decoder, against 18.04M for stock LeWM and 21.55M for Qantara).
LeWM and DINO-WM are retrained the same way on Cube-Double.
The \emph{default recipe} fixes the \(\gamma\!=\!1\) Brownian bridge
on the state axis, loss balance \(\lambda_z\!=\!3, \lambda_a\!=\!1\),
and the inference defaults of \S\ref{sec:method-inference} (CEM
\(K\!=\!4\), BC and video--inverse \(K\!=\!2\)); 10 epochs
correspond to \(\approx\!50\)k optimiser steps at \(B\!=\!128\).

\section{Monotone \texorpdfstring{$\tau$}{tau}-chain on the noisy suffix}
\label{app:tauchain}

For each row, the per-block scalar \(\tau\) on the noisy suffix is
built as a cumulative product of i.i.d.\ uniforms,
\begin{equation}
\tau[t] \;=\; \prod_{i\le t} u_i,\qquad u_i \sim \mathcal U[0,1],
\label{eq:tau-chain}
\end{equation}
monotone-decreasing in the block index. The first noisy block
(immediately after the clean prefix) carries an unbiased
\(\tau\sim\mathcal U[0,1]\); deeper blocks are stochastically noisier
as the product accumulates additional uniforms below \(1\). Together
with the per-row clean-prefix length, the schedule places training
mass on every \((\tau^a,\tau^z)\) pair an autoregressive rollout can
reach. Every noisy block \(s\!\ge\!t\) of Eq.~\ref{eq:sequence} forms
its bridge input (Eq.~\ref{eq:bridge}) from the clean dataset pair
\((z_s, z_{s+1})\), and the state head's residual
(Eq.~\ref{eq:head-z}) is taken on the same clean \(z_s\); for
\(s\!>\!t\) the clean \(z_s\) enters only through the corrupted input
and the residual and is never supplied as a context token.

\paragraph{Empirical role.}
Replacing the cumulative-product chain with i.i.d.\ per-block draws
leaves every per-environment SR within \(\pm 2\,\)SR of the cumprod
chain at the 5-mode reference recipe (single-seed), but destabilises
Push-T training
under reduced mode sets: at the 3-mode
\{\texttt{forward}, \texttt{inverse}, \texttt{policy}\} variant of
the prior \(\lambda_z\!=\!1\) reference recipe, the multi-seed
Push-T mean drops from 84.4\,\(\pm\)\,0.9 (cumprod) to
75.1\,\(\pm\)\,16.6 (i.i.d.; one of three train seeds collapses
Push-T to 53.2\,SR), with the other three environments unchanged
within seed noise. The cumprod chain matches the rolling-denoise
structure an autoregressive rollout induces; removing the inductive
bias is invisible at the over-parameterised 5-mode design but breaks
Push-T under sparser training-mode coverage. We retain the chain as
a safety net.

\section{Brownian \texorpdfstring{$\gamma$}{gamma} sweep, full table}
\label{app:gamma-sweep-full}
Table~\ref{tab:gamma-sweep-full} gives the full Brownian-bridge
\(\gamma\) sweep summarised in \S\ref{sec:exp-robustness}.

\begin{table}[ht]
\small
\centering
\caption{\textbf{Brownian-bridge \(\gamma\) sweep at the default
recipe} (multi-seed, \(K\!=\!1\)). \(\gamma\!=\!1\) is the Albergo et
al.\ stochastic-interpolant canonical \citep{albergo2023stochastic};
\(\gamma\!=\!\sqrt 2\) is the Schr\"odinger-bridge canonical
\citep{liu2023ipsb}; the two tie on suite average. The off-canonical
\(\gamma\!\in\!\{0, 0.5, 2\}\) collapse 1--2 of 3 Push-T seeds and
\(\gamma\!=\!2\) additionally collapses one Two-Room seed.
The \(\gamma\!=\!1\) row is the training-time evaluation of the
Table~\ref{tab:lewm-main} checkpoints; their re-evaluation in
Table~\ref{tab:k-sweep-cem} reads 89.1 on Push-T.
\(^{\ddagger}\)mean masks seed-level collapse.}
\label{tab:gamma-sweep-full}
\begin{tabular}{lcccc}
\toprule
\(\gamma\) & Push-T & Two-Room & Cube & Reacher \\
\midrule
0\(^{\ddagger}\)                              & 31.5\,\(\pm\)\,35.5 & 99.6\,\(\pm\)\,0.0 & 92.5\,\(\pm\)\,0.7 & 78.3\,\(\pm\)\,2.7 \\
0.5\(^{\ddagger}\)                            & 60.1\,\(\pm\)\,39.7 & 99.9\,\(\pm\)\,0.2 & 93.6\,\(\pm\)\,0.6 & 78.8\,\(\pm\)\,1.0 \\
\textbf{1} (Albergo SI canonical, default)    & \textbf{89.3}\,\(\pm\)\,1.4 & \textbf{100.0}\,\(\pm\)\,0.0 & \textbf{93.3}\,\(\pm\)\,0.7 & 80.5\,\(\pm\)\,3.0 \\
\(\sqrt 2\) (Schr\"odinger-bridge canonical)  & 89.3\,\(\pm\)\,0.2 & \textbf{100.0}\,\(\pm\)\,0.0 & 91.9\,\(\pm\)\,1.7 & \textbf{80.8}\,\(\pm\)\,2.6 \\
2\(^{\ddagger}\)                              & 53.6\,\(\pm\)\,36.6 & 71.5\,\(\pm\)\,40.4 & \textbf{93.6}\,\(\pm\)\,0.3 & 79.6\,\(\pm\)\,3.1 \\
\bottomrule
\end{tabular}
\end{table}

\section{Output residual parameterisation}
\label{app:z-delta}

The state-head residual form
\(\hat z_{t+1}\!=\!z_t\!+\!\mathrm{head}_z(h^z_{t+1})\)
(Eq.~\ref{eq:head-z}) pairs with zero-initialised final-layer weights
so the predictor is the identity at initialisation. We ablate the residual by
having the head emit \(\hat z_{t+1}\) directly with default Linear
initialisation, holding the Brownian-bridge input interpolant
\(\tilde z_{t+1}\!=\!(1\!-\!\tau)\,z_t\!+\!\tau\,z_{t+1}\!+\!
\gamma\sqrt{\tau(1\!-\!\tau)}\,\varepsilon_z\) and every other recipe
knob fixed at the default recipe (Table~\ref{tab:z-delta}, row 1 vs row 2).

To complete the factorial we also ablate the input-side
bridge with the residual already removed: the source endpoint
becomes Gaussian noise rather than \(z_t\), so
\(\tilde z_{t+1}\!=\!\tau\,z_{t+1}\!+\!(1-\tau)\,\varepsilon_z\)
(Table~\ref{tab:z-delta}, row 3). Comparing rows 2 and 3 isolates
the bridge contribution at residual\,=\,off; the fourth corner
(bridge\,off, residual\,on) is structurally unreachable in our
implementation since the residual head is gated on the bridge being
active. Restoring the bridge interpolant on top of (residual\,off)
yields a +12.7\,SR lift on Reacher (precision motor control),
a within-noise change on Cube, saturation at 100\,SR on
Two-Room, and a Push-T comparison confounded by the row-2 seed-level
instability noted above. The \(z_t\)-anchored source endpoint is
therefore most informative on Reacher, while the output-side
residual stabilises Push-T and Cube training.

\begin{table}[ht]
\small
\centering
\caption{\textbf{State-head and bridge structural ablations,
multi-seed, all rows at \(K\!=\!1\).} The output-side residual head
(row 1 vs row 2) is load-bearing: removing it costs 5.6\,SR on suite
average and 15.2\,SR on Push-T with one of three train seeds
collapsing. The input-side bridge interpolant (row 2 vs row 3) is
selectively load-bearing on Reacher: restoring it
yields \(+12.7\,\)SR there, within-noise elsewhere. The two
priors are complementary, not redundant.}
\label{tab:z-delta}
\setlength{\tabcolsep}{4pt}
\begin{tabular}{llcccc}
\toprule
bridge & residual & Push-T & Two-Room & Cube & Reacher \\
\midrule
on & on (zero-init; default) & \textbf{89.3}\,\(\pm\)\,1.4 & 100.0\,\(\pm\)\,0.0 & \textbf{93.3}\,\(\pm\)\,0.7 & \textbf{80.5}\,\(\pm\)\,3.0 \\
on & off (default Linear init)             & 74.1\,\(\pm\)\,12.3          & 100.0\,\(\pm\)\,0.0 & 86.3\,\(\pm\)\,1.3          & 80.4\,\(\pm\)\,2.4          \\
off & off (default Linear init)             & 86.5\,\(\pm\)\,0.5           & 100.0\,\(\pm\)\,0.0          & 87.1\,\(\pm\)\,0.8          & 67.7\,\(\pm\)\,1.0          \\
\bottomrule
\end{tabular}
\end{table}

\section{CEM and MPC configuration}
\label{app:cem-mpc-config}

We inherit the CEM and receding-horizon MPC setup from
LeWM~\citep{maes2025lewm}, which in turn follows
DINO-WM~\citep{zhou2024dinowm}, and plan every method trained in our
pipeline with these same settings. CEM uses \(N\!=\!300\) Gaussian
candidates per iteration, \(30\) refit iterations on Push-T and
\(10\) elsewhere, top-\(30\) elites, and an initial sampling variance of \(1\). The planning
horizon is \(H\!=\!5\); under a frame-skip of \(5\) (action-block
repetition), each planned action commands \(5\) environment
steps, so one plan covers \(25\) environment steps. We use a
receding horizon equal to the planning horizon: the entire
elite-mean plan is executed before replanning. All other settings
are identical across the four environments.

\section{CEM denoising-step sweep}
\label{app:k-sweep-cem}

We sweep the state-axis predictor recursion step count
\(K\!\in\!\{1,2,4,8\}\) on the trained Qantara checkpoints
(\S\ref{sec:exp-main}; Table~\ref{tab:k-sweep-cem}). On the
\texttt{forward} edge the recursion starts from the bridge source
\(\tilde z^{(0)}_{t+1}\!=\!z_t\) and makes \(K\) predictor calls at
\(\tau^z_k\!=\!k/K\), \(k\!=\!0,\ldots,K\!-\!1\). Call \(k\) reads
\(\tilde z^{(k)}_{t+1}\) and returns the endpoint estimate
\(\hat z^{(k)}_{t+1}\!=\!z_t\!+\!\mathrm{head}_z(h^z_{t+1})\)
(Eq.~\ref{eq:head-z}), which is re-projected onto the bridge with
fresh noise \(\xi_k\!\sim\!\mathcal N(0,I)\),
\begin{equation}
\tilde z^{(k+1)}_{t+1} \;=\; (1-\tau^z_{k+1})\, z_t \;+\;
\tau^z_{k+1}\, \hat z^{(k)}_{t+1} \;+\;
\gamma\sqrt{\tau^z_{k+1}(1-\tau^z_{k+1})}\,\xi_k,
\label{eq:z-recursion}
\end{equation}
so each call's input lies on the training-time bridge marginal. The
noise term vanishes at \(\tau^z_K\!=\!1\), giving
\(\hat z_{t+1}\!=\!\tilde z^{(K)}_{t+1}\!=\!\hat z^{(K-1)}_{t+1}\); at
\(K\!=\!1\) the recursion is the deterministic single call. The
\texttt{video} factor of the video--inverse composition is this single
call with the action slot at Gaussian noise. The state head is trained by
endpoint regression under the bridge corruption (Eq.~\ref{eq:loss-z}),
and the bridge enters inference through Eq.~\eqref{eq:z-recursion}:
each call returns an endpoint estimate, which the recursion maps back
onto the bridge at the next \(\tau^z\), as endpoint-predicting
diffusion samplers re-noise their clean estimate. We report the same sweep for the
\(\gamma\!=\!\sqrt 2\) checkpoints of
Table~\ref{tab:gamma-sweep-full} for comparison.

\begin{table}[ht]
\small
\centering
\caption{\textbf{CEM predictor recursion step count \(K\) at the
trained Qantara checkpoints, multi-seed.} \(\gamma\!=\!1\), \(K\!=\!4\)
is the default (Table~\ref{tab:lewm-main}). \(K\!=\!8\) regresses on
Push-T at both \(\gamma\) values; \(\gamma\!=\!\sqrt 2\) ties
\(\gamma\!=\!1\) on suite average and is dominated at \(K\!=\!4\).}
\label{tab:k-sweep-cem}
\begin{tabular}{llcccc}
\toprule
\(\gamma\) & \(K\) & Push-T & Two-Room & Cube & Reacher \\
\midrule
1            & 1 & 89.1\,\(\pm\)\,1.3 & 100.0\,\(\pm\)\,0.0 & 93.3\,\(\pm\)\,0.7 & 80.5\,\(\pm\)\,3.0 \\
1            & 2 & 88.0\,\(\pm\)\,0.9 & 100.0\,\(\pm\)\,0.0 & 93.3\,\(\pm\)\,2.1 & 80.1\,\(\pm\)\,1.9 \\
\textbf{1}   & \textbf{4} (default) & \textbf{90.1}\,\(\pm\)\,1.1 & 100.0\,\(\pm\)\,0.0 & \textbf{93.7}\,\(\pm\)\,0.7 & \textbf{80.9}\,\(\pm\)\,1.8 \\
1            & 8 & 87.2\,\(\pm\)\,1.4 & 100.0\,\(\pm\)\,0.0 & 93.2\,\(\pm\)\,0.6 & 78.7\,\(\pm\)\,1.9 \\
\midrule
\(\sqrt 2\)  & 1 & 89.3\,\(\pm\)\,0.2 & 100.0\,\(\pm\)\,0.0 & 91.9\,\(\pm\)\,1.6 & 80.8\,\(\pm\)\,2.6 \\
\(\sqrt 2\)  & 2 & 90.4\,\(\pm\)\,2.0 & 100.0\,\(\pm\)\,0.0 & 93.1\,\(\pm\)\,2.2 & 79.3\,\(\pm\)\,0.5 \\
\(\sqrt 2\)  & 4 & 89.7\,\(\pm\)\,1.5 & 100.0\,\(\pm\)\,0.0 & 92.3\,\(\pm\)\,2.8 & 80.0\,\(\pm\)\,2.3 \\
\(\sqrt 2\)  & 8 & 88.9\,\(\pm\)\,1.3 & 100.0\,\(\pm\)\,0.0 & 92.0\,\(\pm\)\,2.7 & 77.3\,\(\pm\)\,3.0 \\
\bottomrule
\end{tabular}
\end{table}

\section{Multi-paradigm decoded rollouts}
\label{app:rollouts}

Starting from one context frame at \(t\!=\!0\), we roll
the predictor forward six steps under three action sources
(Fig.~\ref{fig:rollout}). The \emph{demo} row replays the
demonstration actions recorded for the held-out trajectory and
exercises the forward-dynamics primitive CEM scores inside its
candidate-search loop; the \emph{BC} row autoregressively chains BC
action sampling with the same forward dynamics; the
\emph{video--inv} row chains video--inverse action extraction with
forward dynamics. All three rows share the starting latent and the
forward-dynamics call, differing only in how the action sequence is
chosen; the \(t\!=\!0\) tile (left of the dashed separator) is the
shared encoded context decoded once, and \(t\!=\!1\,\)--\(\,6\) are
decoded predicted latents. Decoded trajectories remain coherent
across the planning horizon; finer details (T-block angle on Push-T,
end-effector orientation on Cube, agent position in Two-Room and
Reacher) drift after a few steps, matching the characterisation of
decoded JEPA latents in \citet[Fig.~7]{maes2025lewm}. The decoder
behind these figures is a convolutional head trained alongside the
model on stop-gradient predicted latents with a pixel MSE loss: no
gradient flows from the decoder into the encoder or the predictor,
every parameter count in the paper excludes the decoder, and no
dispatch runs the decoder. The training times of App.~\ref{app:optim}
include the decoder, which the LeWM reproduction trains as well.

\begin{figure}[ht]
  \centering
  \includegraphics[width=0.95\linewidth]{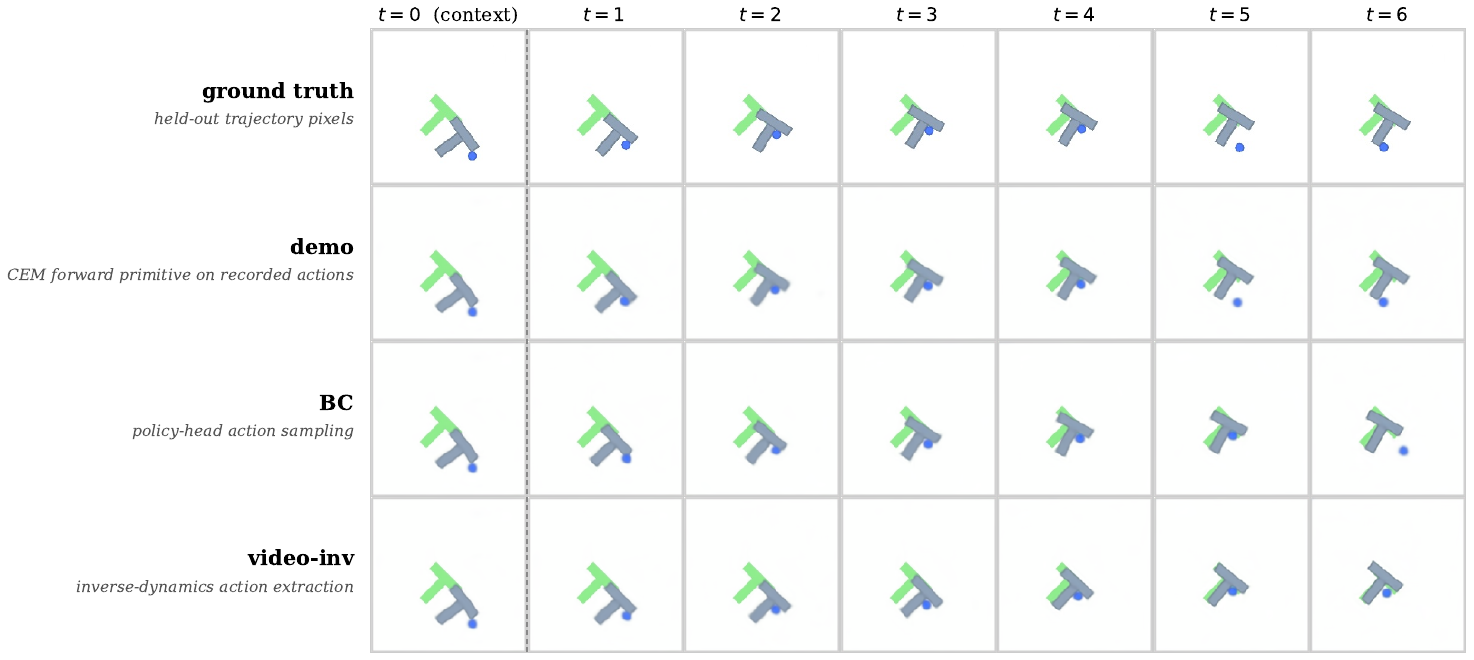}\\[4pt]
  \includegraphics[width=0.95\linewidth]{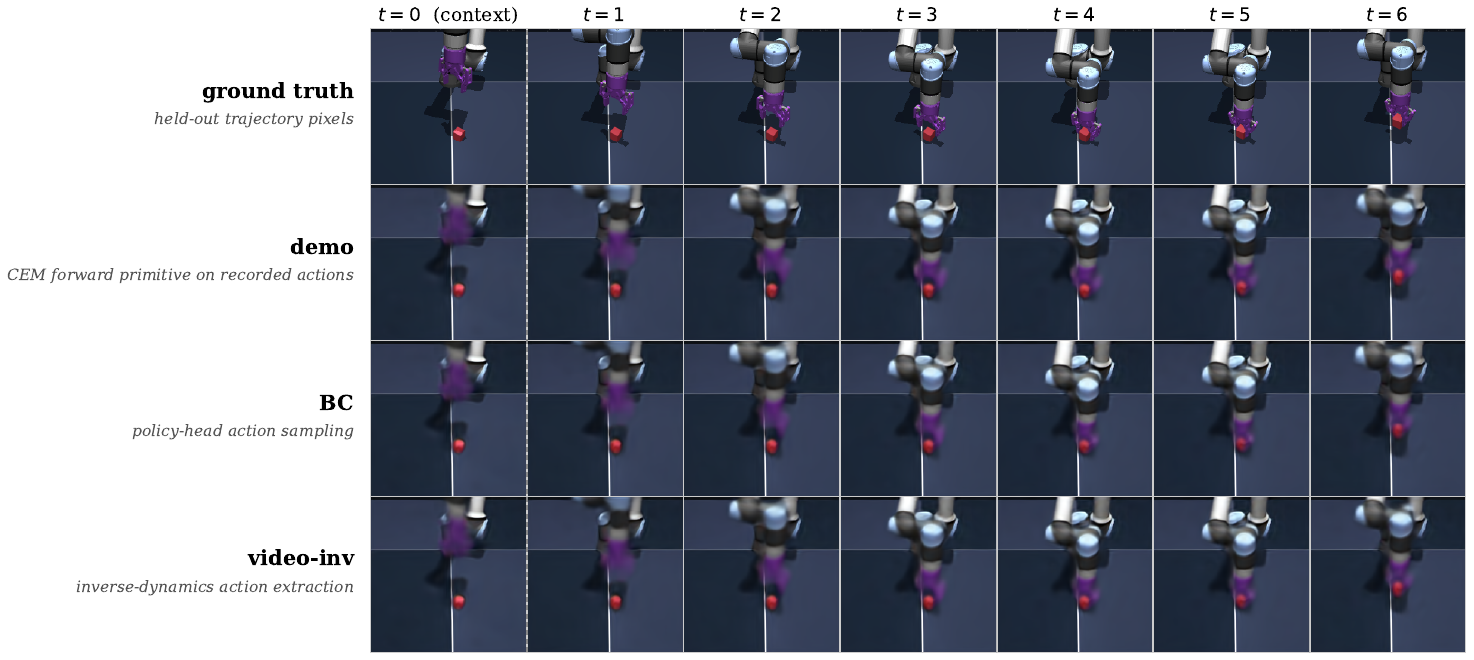}
  \caption{\textbf{Same checkpoint, three action sources, three
  coherent decoded rollouts} (Push-T top, Cube bottom; Two-Room and
  Reacher continued in Fig.~\ref{fig:rollout-cont}). Columns are
  timesteps \(t\!=\!0\,\)--\(\,6\); the dashed separator marks the
  boundary between the encoded context (\(t\!=\!0\)) and the
  open-loop predictor rollout (\(t\!=\!1\,\)--\(\,6\)). Top row:
  ground-truth pixels (decoder not used). The next three rows
  decode the predictor's open-loop latent rollout under \emph{demo}
  (CEM forward primitive on recorded actions), \emph{BC}
  (policy-head action sampling), and \emph{video--inv}
  (inverse-dynamics action extraction) action sources; all three
  rows share the starting latent and the forward-dynamics call.}
  \label{fig:rollout}
\end{figure}

\begin{figure}[ht]
  \ContinuedFloat
  \centering
  \includegraphics[width=0.95\linewidth]{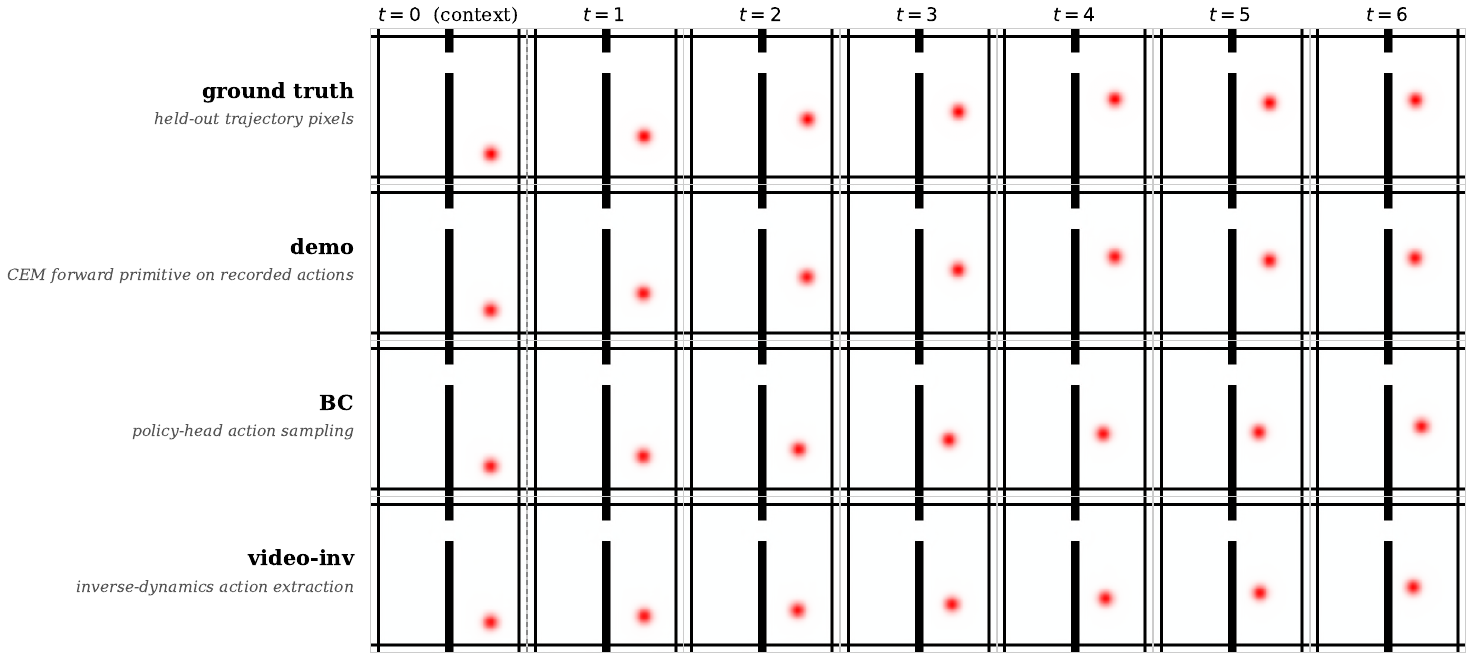}\\[4pt]
  \includegraphics[width=0.95\linewidth]{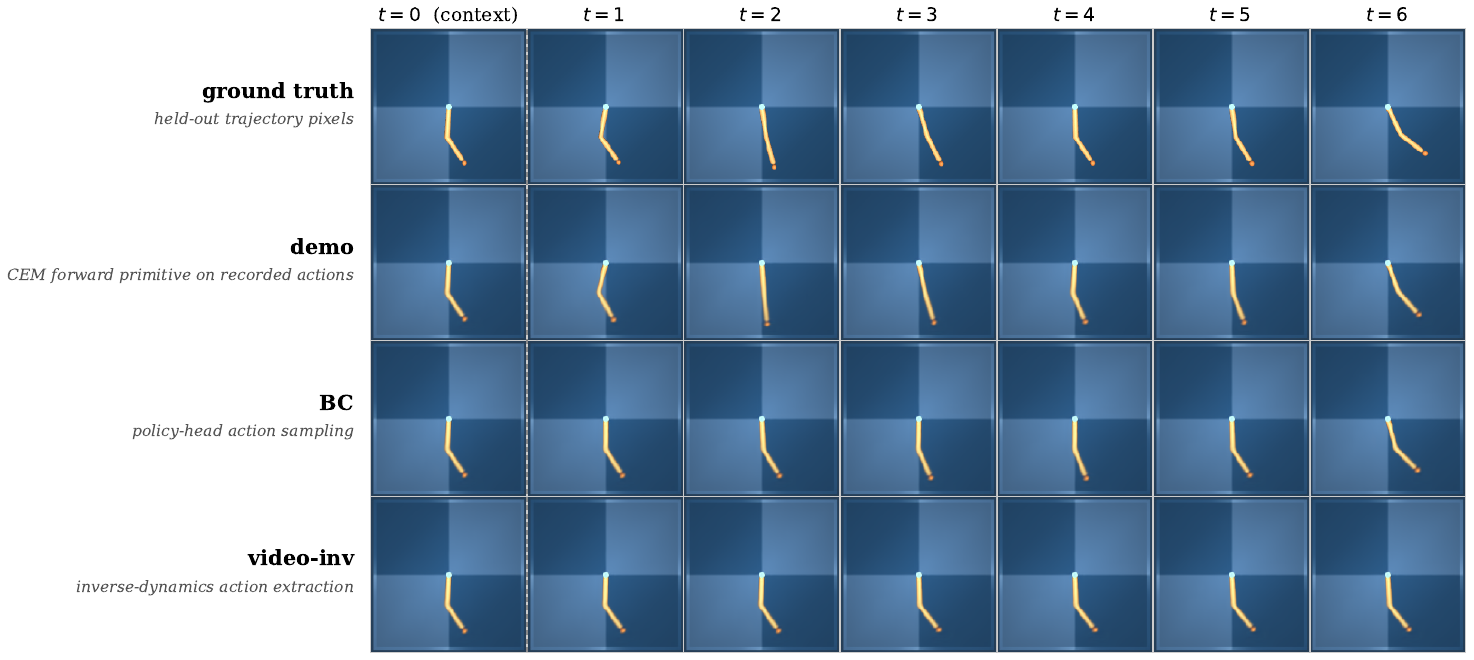}
  \caption{(Continued.) Two-Room (top) and Reacher (bottom) decoded
  rollouts under the same three action sources; layout and column
  semantics as in Fig.~\ref{fig:rollout}.}
  \label{fig:rollout-cont}
\end{figure}

\section{Per-dispatch wall-clock}
\label{app:timing}

Table~\ref{tab:dispatch-time} is the measurement behind the
\(\sim\!15\)--\(65\times\) figure quoted in \S\ref{sec:exp-bc}.
CEM latent planning runs
\(N_{\rm samples}\!\times\!H\!\times\!K\) predictor passes per
refit iteration (\(300\times5\times4=6000\) at the headline settings,
with 30 iterations per plan on Push-T and 10 elsewhere),
while the two goal-blind dispatches need \(N_a\!=\!2\) Euler steps
each. The evaluation calls \texttt{get\_action} once per environment
step, and a dispatch computes only when its action buffer is empty:
CEM once per plan, every 25 environment steps
(App.~\ref{app:cem-mpc-config}), and the goal-blind dispatches once per
5-step frame-skip block. The means in Table~\ref{tab:dispatch-time}
therefore amortise each dispatch over its own replanning cadence, and
one CEM planning call is the worst-case decision latency. The measured ratios span \(14.7\times\) (Reacher
video--inverse) to \(65.4\times\) (Push-T behaviour cloning).

\begin{table}[ht]
\small
\centering
\caption{\textbf{Mean \texttt{get\_action} wall-clock per env-step,
seconds.} One training seed (s11), 5 episodes per environment,
evaluation seed 42, \texttt{num\_envs}\,=\,5, on a single H100, at
the headline solver settings (CEM \(N_{\rm samples}\!=\!300\),
\(n_{\rm steps}\!=\!30\) on Push-T and \(10\) elsewhere, \(K\!=\!4\), horizon 5; BC and
video--inverse \(N_a\!=\!2\)). Times include CUDA
synchronisation. Unlike the success rates in this paper, these are
single-train-seed timings and are reported as an order-of-magnitude
envelope rather than a three-seed mean.}
\label{tab:dispatch-time}
\resizebox{\textwidth}{!}{%
\input{tbl_dispatch_timing}}
\end{table}

\section{Reacher training-budget controls}
\label{app:reacher-budget}
\input{blk_reacher_budget}

\section{Robustness to the remaining knobs}
\label{app:robustness}
\input{blk_robustness}

\section{Capability structure of the joint video--action cluster}
\label{app:va-capability}
Table~\ref{tab:va-capability} lists the capabilities shown in each paper
of the joint video--action cluster (\S\ref{sec:related}).
\input{blk_va_capability}

\section{Trajectory-diffusion and forcing antecedents}
\label{app:diffusion-antecedents}
\input{blk_diffusion_antecedents}

%% file: tbl_dispatch_timing.tex
\begin{tabular}{lccccc}
\toprule
env & CEM latent planning & BC $\tau^z\!=\!0$ Euler & video--inverse & CEM/BC & CEM/vid--inv \\
\midrule
Push-T & 1.166\,$\pm$\,0.003 & 0.018\,$\pm$\,0.000 & 0.019\,$\pm$\,0.001 & 65.4\,$\times$ & 60.1\,$\times$ \\
Two-Room & 0.398\,$\pm$\,0.001 & 0.017\,$\pm$\,0.000 & 0.019\,$\pm$\,0.000 & 22.9\,$\times$ & 20.9\,$\times$ \\
Cube & 0.404\,$\pm$\,0.001 & 0.022\,$\pm$\,0.000 & 0.025\,$\pm$\,0.001 & 18.1\,$\times$ & 16.4\,$\times$ \\
Reacher & 0.403\,$\pm$\,0.003 & 0.024\,$\pm$\,0.004 & 0.027\,$\pm$\,0.003 & 16.7\,$\times$ & 14.7\,$\times$ \\
\bottomrule
\end{tabular}

%% file: blk_reacher_budget.tex
Two measurements place the Reacher gap in the training budget
rather than in the benchmark or the methods. First, the evaluation
instance is consistent with the published one: a random policy
scores 11.2 (mean over 5 eval seeds) under our protocol against the
10.9 reported for the same control by~\citet{maes2025lewm}, as
expected if both draw start--goal pairs from the same corpus, and
unlike what pairs drawn from a converged reaching policy would give
(App.~\ref{app:envs}). Second, the deficit closes at a
larger training budget: retraining the LeWM reproduction for
20 epochs over 5 train seeds lifts Reacher from 64.3 to
\(80.0\,\pm\,0.9\), which is \(1.2\) published standard deviations
below the 86.0 row, while Qantara stays flat (\(79.7\,\pm\,1.8\))
and the two models become indistinguishable there. On a single
train seed the same doubling moves the other three environments by
under \(1\,\)SR, so Reacher is where the 10-epoch budget of
Table~\ref{tab:lewm-main} binds. The Reacher column of
Table~\ref{tab:lewm-main} is therefore a matched-10-epoch
statement for every method in the bottom block, and the
\(+16.6\) Reacher margin of \S\ref{sec:exp-main} does not persist at the longer
budget. PLDM recovers at three train seeds: Reacher
55.1\(\to\)78.4\(\pm\)3.3 over seed-matched 10-epoch cells; at
one seed, Push-T, Two-Room and Cube move \(-4.4\), \(-0.4\),
\(+2.0\,\)SR.

%% file: blk_robustness.tex
Sweeping the state-axis predictor recursion step count
\(K\!\in\!\{1,2,4,8\}\) (\S\ref{sec:method-inference}) on the
trained Qantara checkpoints leaves the suite average within seed
noise (89.8--91.2\,SR; App.~\ref{app:k-sweep-cem}); the
action-axis Euler step count moves Push-T BC SR within
\(2.9\,\)SR across \(K\!\in\!\{1,2,4,8\}\), peaking at \(K\!=\!4\)
with \(+1.1\,\)SR over the \(K\!=\!2\) default we deploy for
latency. The
\(\gamma\!=\!\sqrt 2\) Schr\"odinger-bridge canonical
\citep{liu2023ipsb} ties our \(\gamma\!=\!1\)
stochastic-interpolant canonical \citep{albergo2023stochastic} on
suite average (90.5 vs.\ 90.8\,SR, within seed noise), while
off-canonical \(\gamma\!\in\!\{0, 0.5, 2\}\) are unstable: 1--2
of 3 train seeds collapse on Push-T, and \(\gamma\!=\!2\)
additionally collapses Two-Room on one seed
(App.~\ref{app:gamma-sweep-full}). We retain \(\gamma\!=\!1\).
Finally, the upstream LeWM recipe ships predictor dropout
\(p\!=\!0.1\); ablating to \(p\!=\!0\) at an earlier reference
recipe lifts the suite average from 84.9 to 88.3\,SR (3 train
seeds), and we keep \(p\!=\!0\) as our default.

%% file: blk_va_capability.tex
\begin{table}[ht]
\footnotesize
\centering
\setlength{\tabcolsep}{3pt}
\caption{\textbf{Capability structure of the joint video--action
cluster.} Parameter counts as stated in each paper; ``n/s'': not
stated; ``not shown'': not demonstrated in the paper.
\(\dagger\): ranks a best-of-\(N\) action set by a learned value
function, optimising no test-time goal. \(\S\): re-ranks 100 action
sequences proposed by a separate pretrained policy with a hand-coded
block-to-target reward on predicted frames, optimising no action
sequence against a goal observation. \(\ddagger\): fully
fine-tuned from the 2B Cosmos-Predict2 video model.}
\label{tab:va-capability}
\begin{tabular}{lcccccc}
\toprule
capability & UVA & PAD & UWM & DUST & Cosmos Policy & Qantara \\
\midrule
BC action sampling & yes & yes & yes & yes & yes & yes \\
inverse dynamics & yes & no & yes & not shown & no & yes \\
video / next-obs prediction & yes & yes & yes & yes & yes & yes (latent) \\
goal-conditioned test-time planning & no\(^{\S}\) & no & no & no & no\(^{\dagger}\) & \textbf{yes} \\
prediction space & pixel & pixel & pixel & VLM embed. & pixel & JEPA latent \\
parameters & 0.5B & 0.66B & n/s (DiT-B) & n/s & 2B\(^{\ddagger}\) & \textbf{0.021B} \\
\bottomrule
\end{tabular}
\end{table}

%% file: blk_diffusion_antecedents.tex
Earlier trajectory-diffusion baselines
(Diffuser \citep{janner2022diffuser}, Decision Diffuser
\citep{ajay2023dd}) operate on raw state-action vectors;
Decision Diffuser shares the state-diffusion-plus-inverse-dynamics
factorisation with our video--inverse path, but uses classifier-free
return guidance over multi-step horizons on D4RL/Kuka, whereas
Qantara operates on pixel-encoded JEPA latents at single-step horizon.
Latent Diffusion Planning \citep{xie2025ldp} applies the same
factorisation in a learned image latent space, with a diffusion
planner over future latents and a separate inverse dynamics model on
latent pairs; in Qantara both factors are edges of one checkpoint.
Diffusion Forcing \citep{chen2024dforcing} introduces per-token noise
levels and uses one sequence model over action, reward and
observation tokens both as a policy and, with reward guidance over a
longer lookahead, as a planner; Qantara instead denoises
pixel-encoded JEPA latents, plans by CEM against an encoded goal
observation, and trains on the edges and diagonal of the
\((\tau^a,\tau^z)\) square rather than on i.i.d.\ per-token noise
levels. Follow-ups add clean-prefix conditioning
\citep{zhu2026causalforcing,po2025colds}. Matching the training
distribution of noise levels to the one inference visits is also the
principle behind the Uniform-\(\bar t\) sampler of Spatial Reasoning
Models \citep{wewer2025srm}, which replaces Diffusion Forcing's
i.i.d.\ per-token levels, whose mean concentrates around 0.5;
Qantara applies the principle to the edges and corners of the
\((\tau^a,\tau^z)\) square that the three dispatches query.